\documentclass[pdflatex,sn-mathphys-ay]{sn-jnl}

\usepackage{booktabs}
\usepackage{amssymb}
\usepackage{amsmath}
\usepackage{amsfonts}
\usepackage{array}
\usepackage[title]{appendix}
\usepackage{xcolor}
\usepackage{rotating}
\usepackage{mathrsfs}
\usepackage{url}
\usepackage{etoolbox}
\usepackage{textcomp}
\usepackage{manyfoot}
\usepackage{algorithm}
\usepackage{algorithmicx}
\usepackage{algpseudocode}
\usepackage{listings}
\usepackage{array}
\usepackage{hhline}
\usepackage{multirow}
\usepackage{graphicx}
\usepackage{longtable}
\usepackage{url}
\usepackage{breakurl}

\newcolumntype{P}[1]{>{\centering\arraybackslash}m{#1}}

\begin{document}

\title[Article Title]{A Novel Loss Function for Deep Learning Based Daily Stock Trading System}

\author*[1,3]{\fnm{Ruoyu}\sur{Guo}}\email{guor4@vcu.edu}

\author[2,3]{\fnm{Haochen}\sur{Qiu}}\email{hq238@cam.ac.uk}

\author[3]{\fnm{Xuelun}\sur{Hou}}\email{hou@brandeis.edu}

\affil[1]{\orgdiv{Present address: Department of Mathematics and Applied Mathematics}, \orgname{Virginia Commonwealth University}, \orgaddress{\street{1015 Floyd Ave}, \city{Richmond}, \postcode{23284}, \state{VA}, \country{USA}}}

\affil[2]{\orgdiv{Present address: Department of Mathematics}, \orgname{University of Cambridge}, \orgaddress{\street{Centre for Mathematical Sciences, Wilberforce Rd}, \city{Cambridge}, \postcode{CB3 0WA}, \country{United Kingdom}}}

\affil[3]{\orgdiv{Department of Mathematics}, \orgname{Brandeis University}, \orgaddress{\street{415 South Street}, \city{Waltham}, \postcode{02453}, \state{MA}, \country{USA}}}

\abstract{
Making consistently profitable financial decisions in a continuously evolving and volatile stock market has always been a difficult task. Professionals from different disciplines have developed foundational theories to anticipate price movement and evaluate securities such as the famed Capital Asset Pricing Model (CAPM). In recent years, the role of artificial intelligence (AI) in asset pricing has been growing. Although the black-box nature of deep learning models lacks interpretability, they have continued to solidify their position in the financial industry. We aim to further enhance AI's potential and utility by introducing a return-weighted loss function that will drive top growth while providing the ML models a limited amount of information. Using only publicly accessible stock data (open/close/high/low, trading volume, sector information) and several technical indicators constructed from them, we propose an efficient daily trading system that detects top growth opportunities. Our best models achieve 61.73\% annual return on daily rebalancing with an annualized Sharpe Ratio of 1.18 over 1340 testing days from 2019 to 2024, and 37.61\% annual return with an annualized Sharpe Ratio of 0.97 over 1360 testing days from 2005 to 2010. The main drivers for success, especially independent of any domain knowledge, are the novel return-weighted loss function, the integration of categorical and continuous data, and the ML model architecture. We also demonstrate the superiority of our novel loss function over traditional loss functions via several performance metrics and statistical evidence.
}


\maketitle

\section{Introduction}
\label{sec:Introduction}

Stock price and movement prediction have always been extraordinarily challenging yet heavily sought-after tasks. Before the popularity of artificial intelligence and availability of unforeseen computing power present today, initial stages of our financial understanding consist of the Capital Asset Pricing Model (CAPM) \citep{sharpe1964capital}, the Efficient Market Hypothesis (EMH) \citep{fama1970efficient}, and more. Decades of research following them have witnessed a vast number of articles that build upon these very fundamental concepts, including the 3, 4, and 5-factor models \citep{fama1993common, carhart1997persistence, fama2015five}. In recent years, these theories still form the foundation underlying a significant amount of work such as the Markov Decision Process \citep{park2024novel} and ARIMA \citep{box2015time}, which have adopted more advanced mathematical and statistical techniques.

The two main branches of techniques used to analyze stock prices in order to develop profitable trading strategies are fundamental and technical. The fundamental approach considers both internal factors such as price to earnings (PE), earnings per share (EPS) in financial statements and external factors such as the macroeconomic environment and geopolitical conditions. Analyzing textual data such as news and online posting platforms in the field of Natural Language Processing (NLP) has also grown significantly in popularity in light of the success of Large Language Models (LLMs). There are many models that utilize word embeddings such as GloVe \citep{pennington2014glove, zhang2022transformer}, BERT \citep{devlin2018bert} or more \citep{lin2022factors} to process texts, which can be used as one component of the inputs or for sentiment analysis. The embeddings from LLMs can replace these traditional embedding techniques due to LLM's much stronger language ability \citep{kim2024profitability}. On the other hand, our work falls under the technical side, where we primarily focus on historical price and volume data and the additional features that are constructed from them. To name a few examples utilizing primarily technical indicators, \citet{Hoseinzade2019CNNpred} incorporated information from related markets as the third dimension of the data; \citet{gezici2024deep} converted time-series Exchange-Traded Funds (ETFs) data containing technical indicators into 2-dimensional images and uses vision transformers, image transformers, and Swin to make trading decisions. There were also many successful studies that use a combination of sentiment and technical analysis \citep{picasso2019technical, prachyachuwong2021stock}.

Many technical factors have been found and used over the years, and the market can become increasingly more efficient with respect to them \citep{chen2021open}. As a result, the linear relationship between the realized return and these factors is largely priced in. However, the inherent non-linear relationship among the factors and market movement is not captured easily, so stock market predictions remain a challenging task. To this end, recent advances in deep learning have led to the adoption of models such as Convolutional Neural Networks (CNN), Long Short-Term Memory (LSTM), and attention networks to enhance prediction accuracy \citep{lu2020cnn, luo2024short, zhao2023stock}. While these models have shown promising results, most studies focus primarily on minimizing prediction errors rather than optimizing financial performance \citep{oncharoen2018deep}. Traditional loss functions fail to account for the financial impact of prediction errors, treating all errors equally rather than weighting them based on potential trading risks and returns.

To address this limitation, we propose a novel enhancement to loss functions in the context of non-linear machine learning (ML) models. By incorporating return-weighted loss functions, where prediction errors for stocks with higher absolute returns are penalized more heavily, the model aims to align loss optimization with actual financial impact, especially those with high growth potentials. The highlights and key contributions of the paper are as follows.

\begin{enumerate}
\item We propose a ML-based efficient trading system equipped with a novel return-weighted loss function that is specifically adept at capturing top growth opportunities.
\item We guide the continuous time-series feature data with static categorical sector data by adding the sector categorical embedding cross-sectionally to the feature dimension.
\item We demonstrate the superiority of our novel loss function over traditional loss functions via return and statistical analysis.
\end{enumerate}

The remainder of the paper is organized as follows. We mention related works in Section \ref{sec:Related Works}. In Section \ref{sec:Problem Formulation}, we briefly formulate the objective of the study and the setup we establish to achieve it. Section \ref{sec:Implementation} contains data, model training details, and various analysis to demonstrate the superiority of our proposed novel loss function. We conclude the study with a brief discussion of its limitations and provide several future direction in Section \ref{sec:Conclusions}.

\begin{figure}
\centering
\includegraphics[width=\linewidth]{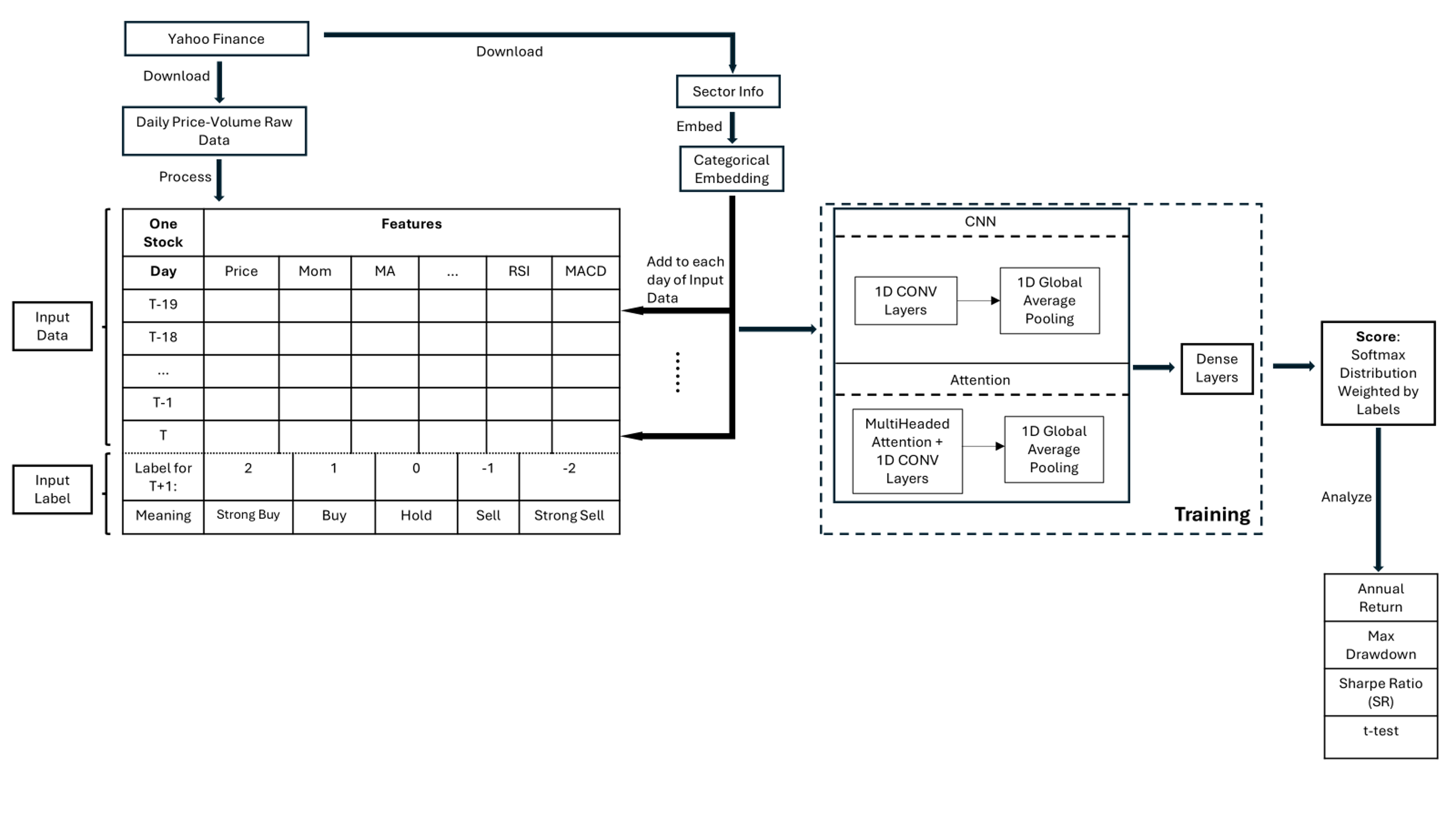}
\caption{Study Pipeline Schematic}
\label{fig:schematic}
\end{figure}

\section{Related Works}
\label{sec:Related Works}

There have been undoubtedly a tremendous amount of literature studying the application of deep learning in financial markets. Early works such as \citep{Dingli2017} tested CNNs on stock direction forecasting, which achieved accuracies of 60–65\%, although the result did not clearly surpass traditional methods like logistic regression or multilayer perceptrons. \citet{Hoseinzade2019CNNpred}  demonstrated that CNNs can help obtain useful features from diverse market variables which were then used to improve high-frequency forecasting on the CSI 300 index, especially under extreme market conditions such as the two time periods we choose in this study as well. Machine learning methods can supplement each other and benefit from other theoretical concepts. For instance, \citet{Mohan2022} combined chaos theory with CNNs to enhance predictive accuracy. \citet{trivedi2025enhancing} converted basic price data (open, close, high, low) into inputs compatible with pre-trained language models so that language models are able to process both textual and numerical information. They later fine-tuned language models to make stock trend predictions. \citet{Zeng2023} introduced a CNN–Transformer hybrid that outperformed ARIMA and DeepAR on intraday S\&P 500 data, using CNNs on short-term features and transformers to model long-term dependencies.

In addition to directional forecasting, other studies focused on linking prediction to trading performance. \citet{cryptoJiangLiang} applied CNNs within a reinforcement learning framework for cryptocurrency portfolio management, directly outputting portfolio weights to maximize returns, but they noted that the Sharpe ratios are not stable. More recently, \citet{Wang2025} proposed a self-adaptive trading strategy that dynamically switches forecasting models and trading types, achieving both higher returns and lower drawdowns in the Hong Kong equity market.

We observe that, however, detailed studies on the loss function that models optimize are largely absent from existing work. Comprehensive surveys \citep{terven2025comprehensive, wang2022comprehensive} pointed out the common loss functions are regression metrics such as mean squared error (MSE) and mean absolute percent error (MAPE) and classification metrics such as cross entropy (CE). Each of these aforementioned loss functions have a vast number of possible variations adapting to specific tasks. Our proposed loss function is like a weighted cross entropy loss, with the weight on the granular level of each data point. However, typical weighted loss functions are typically used on the whole classes of labels to alleviate class-imbalance issues \citep{lin2017focal}.

In the financial domain specifically, expository works \citep{ashtiani2023news, patel2023deep} on the role of ML in finance indicate that the choices of loss functions (and evaluation metrics) in the current deep learning literature are indeed overwhelmingly repetitive. They each have their advantages in specific tasks, but a potential drawback of them is that they treat each prediction equally and then take some form of average. Even when they do well across the entire dataset, they may not be adept at practical financial applications such as finding the small number of top growth opportunities. Some recent research has explored improvements to loss functions in stock prediction. A generalized loss function is proposed that dynamically adjusts the cost of prediction errors based on data difficulty, achieving higher returns despite similar accuracy levels \citep{zhao2024generalized}. Combining the advantages of both classification and regression metrics, we propose a novel loss function with a weighting scheme that prioritizes the model's attention on those with high growth potential and high risk for steep loss. Specifically, the novel loss function is
\[
\mathrm{loss}(y_{\text{true}}, y_{\text{pred}}) = \mathrm{CE}(y_{\text{true}}, y_{\text{pred}}) * |r_{\text{cap}}|,
\]
where CE is cross-entropy and the weight $r_{\text{cap}}$ is the capped version of daily percentage return introduced in Section \ref{sec:First Step of Loss Function - Daily Return}. We implement the novel loss function in ML-based models, where stocks with higher percentage gains and losses are penalized more heavily during training. This allows us to align loss optimization with actual financial impact, leading to improved decision-making for traders and investors. The proposed method is evaluated using real-world stock market data, comparing its performance against linear regression models and traditional loss functions.

\section{Study Design based on Novel Loss Function}

\subsection{Problem Formulation and Strategy Overview}
\label{sec:Problem Formulation}
We propose a novel loss function and use it to develop trading systems that produce a score for every stock considered on a daily basis. These scores are calculated after the end of a trading day $T$ to rank all stocks and recommend the top stocks to invest in on the next trading day $T+1$. To calculate the score for a stock, we input into the model the following 2-dimensional matrix
\[
X = (x_{ij}), i=1,\dots, m, j=1,\dots,n
\]
of size $m\times n$, where $n$ is the number of technical features we consider, and $m$ represents the number of days to which the value of the feature goes back. Namely, we consider the time series of each feature from day $T-m+1$ to day $T$ when predicting the action for day $T+1$.

The main component of our trading systems is a neural network consisting of various structures taking technical features as inputs. There are five initial price and volume features that are easily and publicly accessible (open, close, high, low, trading volume). We compute 27 additional features related to trading volume, momentum, volatility, and trend that are computed from the five initial features (See Table \ref{tab:all features} for some of their description in the Appendix). Disregarding the four initial price features since they are unlikely closely correlated with stock return, we obtain $n=28$ features as the input of the model.

The neural networks are the first step in calculating the score that determines our investment strategy. Specifically, we train neural networks guided by a novel loss function to map the input data $X$ to a probability distribution $y_{pred}$ of 5 labels: Strong Sell, Sell, Hold, Buy, and Strong Buy, respectively. In the second step, the score is calculated as the dot product between the predicted distribution $y_{pred}$ and the vector $[-2,-1,0,1,2]$. We rank each stock after combining the scores from three independent models and invest in the highest ranked stocks equally.

\subsubsection{First Step of Loss Function - Daily Return}
\label{sec:First Step of Loss Function - Daily Return}

The practical implementation of our novel return-weighted loss function begins with the calculation of daily returns. The daily return $r_{d, T}$ (or $r_d$ if the day $T$ is implicit) of a stock is the percentage change from the opening price $p_{open, T+1}$ of day $T+1$ to the opening price $p_{open, T+2}$ of day $T+2$. We calculate $r_{d,T}$ as 

\begin{equation}
\label{eq:daily return}
r_{d, T} = \frac{p_{open, T+2} - p_{open, T+1}}{p_{open, T+1}},
\end{equation}
which is then used to assign the following labels in the data as follows
\begin{itemize}
\item $r_{d,T} \geq 3\%$ - Strong Buy - $y_{true} = [0,0,0,0,1]$,
\item $1\% < r_{d,T} < 3\%$ - Buy - $y_{true} =[0,0,0,1,0]$,
\item $-1\% < r_{d,T} \leq 1\%$ - Hold - $y_{true} =[0,0,1,0,0]$,
\item $-3\% < r_{d,T} \leq -1\%$ - Sell - $y_{true} = [0,1,0,0,0]$,
\item $r_{d,T} \leq -3\%$ - Strong Sell - $y_{true} = [1,0,0,0,0]$.
\end{itemize}

The data distribution after the labeling is (7.43\%, 18.50\%, 46.73\%, 19.45\%, 7.88\%) in the 2005-2010 period and (7.79\%, 17.61\%, 47.83\%, 18.92\%, 7.85\%) in the 2019-2024 period. It signals a slightly positive skew in the data which can also skew the model predictions to be slightly more positive, which is discussed later in Section \ref{sec:Net Asset Value}. The daily return $r_{d,T}$ is the objective for the three baseline linear regression models and another baseline CNN with MSE loss introduced in Section \ref{sec:Baseline Methods}. In the formulation of the new loss function, $r_{d,T}$ serves as the weight for the cross-entropy loss that we describe in Section \ref{sec:Novel Loss Function}.

\subsection{Novel Loss Function}
\label{sec:Novel Loss Function}

Both the data and the loss function play a significant role in ML model's performance. While data provide the resources for model training, the loss function that an ML model minimizes is a guide that directly impacts the performance and (downstream) applications of the model. Typical loss functions include Mean Squared Error (MSE) or its variants in the case of asset pricing and cross-entropy (CE) in the case of classifications. There are three notable considerations in using the very widely used MSE loss in the current study as follows.
\begin{enumerate}
\item MSE loss is typically associated with asset pricing tasks, which are extremely challenging by nature due to noise, unpredictable market conditions, sudden news releases, events, and more. One could argue that detecting bullish and bearish signals, despite difficult in its own right, is the more manageable task compared to actual future price predictions.
\item The presence of outliers can skew the model prediction significantly. This issue can sometimes be counteracted by increasing the size of the data, but that is not always possible. It can also be combated by winsorizing the model objective, similar to the role of $r_{\text{cap}}$ defined in \eqref{eq:r_cap}.
\item MSE loss treats all data equally. While it may be better at stock ranking when used on good quality data and empirically sound model architecture, it is not designed to detect top growth opportunities.
\end{enumerate}

The other common loss function cross-entropy \citep{shannon1948mathematical} is a measure of the difference between two probability distributions, so it is excellent at classification tasks. The cross-entropy of a predicted distribution $q=(q_1,\dots, q_n)$ relative to the true distribution $p=(p_1,\dots,p_n)$ is
\begin{equation}
\label{eq:cross entropy}
\mathrm{CE}(p, q) = -\sum_{i=1}^n p_i \log q_i,
\end{equation}
where the $\log$ has base $e$. In the current study where the true label $p$ is a one-hot vector, i.e., exactly one of $p_i$ is 1 and the others are 0, the sum \eqref{eq:cross entropy} only has one nonzero summand $p_i\log q_i=\log q_i$.

The cross-entropy loss can treat incorrect classifications equally even when they have vastly different financial implications. For example, mislabeling Strong Sell and mislabeling Hold could lead to the same cross-entropy loss. Suppose there are two true labels Strong Sell (next day return $r_1=-5\%$, label $y_1=[1,0,0,0,0]$) and Hold (next day return $r_2=0.5\%$, label $y_2=[0,0,1,0,0]$), and that there are two predictions $\hat{y_1}=[0.2, 0.1, 0.1, 0.1, 0.5]$ and $\hat{y_2}=[0.1, 0.1, 0.2, 0.5, 0.1]$. By the definition of cross-entropy \eqref{eq:cross entropy}, $\text{CE}(y_1, \hat{y_1}) = \text{CE}(y_2, \hat{y_2}) = -\log (0.2)$, but in practice, these two mistakes are not considered the same in terms of their financial consequence. Therefore, we propose a weighting scheme for the cross-entropy loss function that naturally prioritizes stocks with large price movement so that the model will learn to focus on predicting those stocks correctly. To protect the models from being overly skewed by large values of daily percentage return $|r_d|$, we use a capped version of the daily return $r_d$ as the weight. The definition of the capped version $r_{\text{cap}}$ with cutoff at $50\%$ is

\begin{equation}
\label{eq:r_cap}
r_{\text{cap}} = \begin{cases} 
r_d & \text{if } |r_d| \leq 0.5, \\
0.5 & \text{if } |r_d| > 0.5.
\end{cases}
\end{equation}

Our novel return-weighted loss function is 

\begin{equation}
\label{eq:loss function}
\mathrm{loss}(y_{\text{true}}, y_{\text{pred}}) = \mathrm{CE}(y_{\text{true}}, y_{\text{pred}}) * |r_{\text{cap}}|,
\end{equation}
where $y_{\text{true}}$ and $y_{\text{pred}}$ are probability distributions over five labels.

Note that in the example above, our new loss has that $\text{loss}(y_1,\hat{y_1})=-0.05\log(0.2)$, which is 10 times as much as $\text{loss}(y_2,\hat{y_2})=-0.005\log(0.2)$. This aligns with our understanding that incorrectly classifying Strong Sell stocks is worse than incorrectly classifying Hold stocks. In fact, since the stocks labeled as Strong Sell or Strong Buy in the dataset both have their absolute daily percentage change greater than 3\%, they always contribute to the loss function more significantly than those with the other three labels. Therefore, by minimizing loss during training, the models will identify and prioritize stocks with immediate breakout and breakdown potentials. Most stocks do not move significantly on a daily basis, and they do not increase the loss as much even when they are classified incorrectly. This leads to better detection for top growth opportunities but could lead to slightly worse overall ranking ability which we discuss in Section \ref{sec:Net Asset Value}.

\subsection{Model Architecture}
\label{sec:Model Architecture}

We describe various neural network setups in this section and when applicable, the specific configurations that we use in this study.

\textbf{Multi-Layered Perceptrons (MLPs):} Also called Deep Neural Networks (DNNs), MLPs are relatively simple supervised machine learning algorithms in terms of data preparation and implementation. Given the versatility of these neural networks in general, they have been deployed in lots of financial literature and have achieved great success \citep{gu2020empirical}. On the other hand, one key drawback of DNNs is that they only take 1-dimensional inputs. While it is possible to collapse multi-dimensional data into 1 dimension, DNNs do not have an inherent understanding of higher dimensional data structure. During learning, they might be able to pick up on the dimensionality in later hidden layers, but this is far from guaranteed and is at the cost of resources.

\textbf{Convolutional Neural Networks (CNNs):} CNNs are originally developed for tasks involving images, such as classification \citep{lecun1998gradient}, segmentation \citep{ronneberger2015u} and more. Colored images are naturally 3-dimensional data: their three RGB channels being one dimension, and the height and width being the other two dimensions. For each color channel in the input data, convolutional layers apply a sliding window of fixed size $l$ by $w$ (also called kernel) and compute a linear combination of the data values in each window. The coefficients (also called weights) used in the combinations are trainable and are shared across all kernel positions within each input channel. Convolutions use separate weights for different input channels. One output channel is obtained by summing the results across input channels at each position. If there are multiple output channels, this process is repeated with different sets of weights for each output channel. One premise of using convolutions is that data points that are in the same kernel interact with each other in some way. In our study, we perform 1D convolutions on the time dimension, so the kernel has one dimension of size $l$, and this setup helps the model find patterns in each feature's trend over time. Since there may not be a reasonable way to order the features such that all nearby features interact with each other, we do not convolve the features.

\textbf{Attention:} Introduced by Vaswani et al. (\citeyear{vaswani2017attention}), the attention mechanism excels at language related and generative tasks. To name a few popular applications, attention is a crucial component in large language models and diffusion models. Given a sequential input such as a sentence or a time series of stock data in our case, we first construct contextual questions (Q) and keys (K) as attempted answers. The interactions between the questions and keys weight the values (V) to produce the output of one attention head. Multiple independent attention heads can be constructed in parallel forming Multi-Head Attentions (MHA), and their outputs are concatenated before taken to subsequent dense layers. The MHA layer and its subsequent layers make up one encoder block.
Residual connections \citep{he2016deep} are used among layers in the encoder block to speed up data passage. There are typically multiple consecutive encoder blocks in practice.

\textbf{Sector Embedding:} We embed a trainable one-hot vector calculated from the 12 sector dimensions into $n=28$ dimensions. The embedding is added to the third (feature) dimension of the input data $X$ of shape (None, $m$, $n$), and the embedding matrix is trained together with all other parameters. The way in which the sector embedding is added to the input data resembles how positional embeddings are used in transformer models \citep{vaswani2017attention}. In this particular application in stock selection, there are several reasons for why we need to use sector information to guide the technical features. The features we constructed for each stock should be interpreted differently in different sectors. Certain features have varying levels of impact depending on the sector they are in, and some features exhibit different levels of longevity and periodicity. Moreover, the embedding matrix only contributes $12\times 28=336$ additional parameters, serving as an economical method to integrate categorical data with continuous data.

\begin{figure}[htp]
\centering
\includegraphics[width=\linewidth]{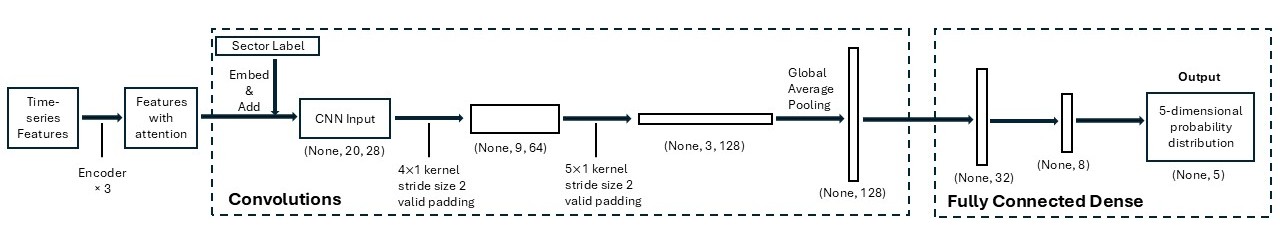}
\caption{CNN and attention model architecture equipped with novel loss function}
\label{fig:architecture}
\end{figure}

For non-ML models, no sector embedding is added. For CNN models, after the sector embedding is added to the input data, the result goes through several layers of 1D convolutions without padding on the time dimension, thereby reducing the time dimension and increasing the feature dimension of the data. We eventually remove the time dimension through a global average pooling layer before passing the result to a series of dense layers. For attention-based models, the input data first go through three encoder blocks whose output has the same dimension as the input. Then, sector embedding is added as in the CNN models and the subsequent structures are the same. See Figure \ref{fig:architecture} for an illustration.

\subsection{Baseline Methods}
\label{sec:Baseline Methods}
To obtain baseline performance for comparison, we choose several sets of features to conduct linear regression on. Since linear regression methods do not directly fit to the same probability distribution objective of the ML models, the objective of linear regression is the daily return $r_d$ \eqref{eq:daily return}. The sets of chosen features that we perform regression on are
\begin{enumerate}
\item 3-day momentum, number of shares traded;
\item 3-day momentum, relative strength index;
\item 3-day momentum, 50-day moving average;
\item 2-day momentum, 5-day momentum;
\item a subset of 12 non-technical features;
\item all 28 features.
\end{enumerate}

We perform ordinary least squares (OLS), ridge and lasso with cross validation on these six sets of features. The data used for linear regression are identical to those for the ML models.

\section{Implementation, Investment Simulation and Analysis}
\label{sec:Implementation}

\subsection{Data Management}
\label{sec:Data Management}

This subsection is dedicated to presenting the data pipeline in the study. We obtain our data from the publicly available Yahoo Finance database. Given that our current study only requires basic daily information including price (open, close, high, low), trading volume, and sector information, other open source databases can be used as well. All stocks in the dataset are from major US stock exchanges, such as S\&P 500, New York Stock Exchange (NYSE), and Nasdaq. The data we use in this study are publicly available at \href{https://data.mendeley.com/datasets/czwwfgcgz7/1}{Mendeley Data}.

\subsubsection{Data Sourcing and Processing}
\label{sec:Data Sourcing}

We download daily stock price (open, close, high, low) data, volume (number of shares traded) data, and sector information from Yahoo Finance using its Python API. There are 11 sectors in the dataset, and each stock's sector information is stored as a one-hot vector in 12 dimensions, where the extra dimension is for those without sector information in Yahoo Finance (such as ETFs). We intentionally choose two time periods 2006-2010 and 2019-2024 to include the 2008 financial crisis and its aftermath, the 2020 COVID pandemic, and the heightened fluctuations observed in the market starting around 2022 due to high inflation, fear of recession, and geopolitical factors. We process stocks by removing microcap stocks and those with low dollar trading volume. This results in 924 stocks in the 2006-2010 period with 1360 trading days for backtest and 2152 stocks in the 2019-2024 period with 1340 trading days for backtest.

\subsubsection{Feature Engineering}
\label{sec:Feature Engineering}

We construct two sets of features from the price and volume data obtained in Section \ref{sec:Data Sourcing}. The first set contains more commonly known features such as momentum and basic statistics of the data such as moving averages and standard deviation. The second set of features are more technical and contains a subset of the indicators obtained from the Technical Analysis (TA) library \citep{TALib}. Since our simulated trades all occur at market open, all features are calculated with the \textit{opening} prices instead of \textit{closing} prices by default. In order to maintain the efficiency of resources, we choose features broadly first and then limit the number of features to only 28 in total by removing those that are highly correlated with the existing features. Recall that we establish in Section \ref{sec:Problem Formulation} that the input shape is (None, $m$, $n$), where the first dimension is the mini-batch size, $m=20$ is the length of time-series, and $n=28$ is the number of features. Some technical features and their brief descriptions are in Table \ref{tab:all features} in the Appendix.

\subsubsection{Standardization and Training Setup}
\label{sec:Standardization}
We employ a fixed size moving window for training, validation, and testing. The training/validation set spans 200 days, and the test set contains 20 days immediately after. We standardize data across the time dimension to fully take advantage of the time-series nature of stock data. The standardization set contains the data 200 days immediately prior to the training/validation set. For example, suppose we fix a stock $A$ and compute for each feature $j$ its mean $\mu(x_j)$ and standard deviation $\sigma(x_j)$ over the 200 days in the standardization set. Then given the raw value $x_{ij}$ of a feature $j$ on day $i$ of stock $A$ in the training/validation and test set, the standardized value of $x_{ij}$ is

\begin{equation}
\label{eq:standardization}
\displaystyle
x_{ij}^{\text{st}} = \frac{x_{ij} - \mu(x_j)}{\sigma(x_j)}.
\end{equation}

\begin{figure}[htp]
\centering
\includegraphics[width=\linewidth]{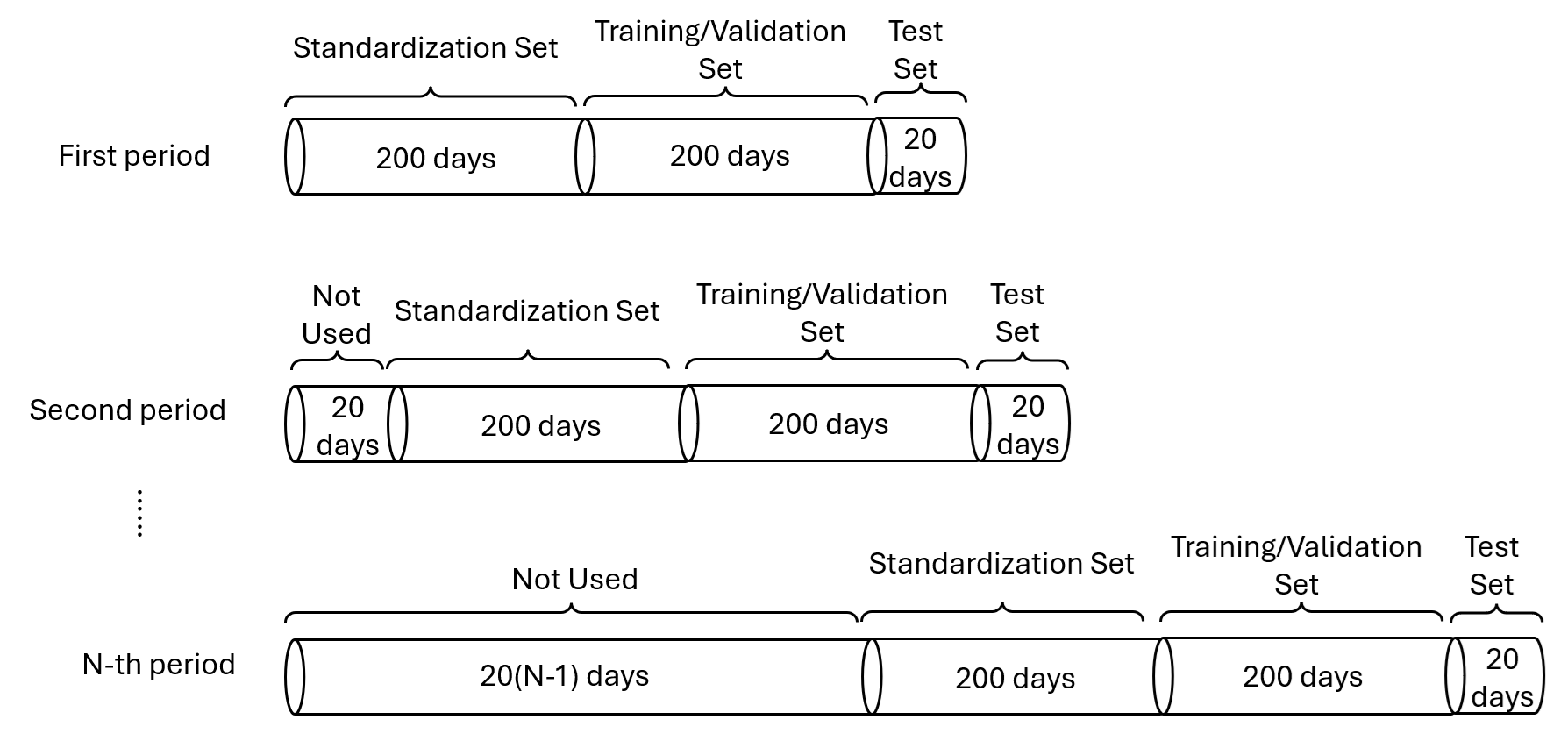}
\caption{Training, Validation, Testing Data Split for Each Backtest Period}
\label{fig:data split}
\end{figure}

In doing so, the first training/validation set starts on day 201. We use the same standardization set for training, validation, and testing to ensure the similarity of their data distributions. Every 20 trading days in the test set form a testing period. For the next testing period, we move all sets 20 days to the future and maintain their size. See Figure \ref{fig:data split} for a visualization.

There has been a lot of work where cross-sectional data are used for predictions, such as \citet{gu2020empirical}, where instead of standardization, they perform normalization and map the cross-sectional feature ranks into a predetermined interval such as $[-1, 1]$. In our case, $\mu(x)$ and $\sigma(x)$ are the mean and standard deviation of the feature values of the \textit{same} stock over a certain amount of historical data, which prioritizes the time-series nature of the data in this particular case.

\subsection{Hyperparameter and Other Settings}
\label{sec:Hyperparameter Setting}

For CNN models equipped with the novel loss function, we set the training patience to be 20 epochs, initial learning rate to be 0.01 with the Adam optimizer \citep{kingma2017adam}, plateau patience to be 5 so that the learning rate is halved if the validation metric does not improve after 5 epochs until either the training patience runs out or the learning rate reaches the minimum learning rate 0.001, after which the learning rate no longer decreases. We apply batch normalization \citep{ioffe2015batch} and leaky ReLU activation after each convolution and dense layers, in that order. To reduce overfitting, we also apply dropout \citep{srivastava2014dropout} with 35\% rate after the activation function of both CNN and dense layers. Attention models have 15\% dropout rate for the layers in the encoder and 35\% in other layers that coincide with CNN. The starting and minimum learning rates for attention model are also adjusted to 0.0025 and 0.0001 for better convergence, respectively. Detailed implementations of these models are available in \href{https://github.com/Tony-Guo-1/daily_trading_strategy}{GitHub}.

In the second testing period and onward, we continue with a sliding window of fixed size 20 for retraining and testing (Figure \ref{fig:data split}). During retraining, we set the learning rate back to its initial rate and apply the same learning rate schedule. This allows the models to explore more of the parameter space and find better local minima during training. The CNN model equipped with the novel loss function has only 53,280 trainable parameters and takes approximately 2 seconds per epoch when running on NVIDIA GeForce RTX 4060 Laptop GPU with 16 GB GPU memory. The attention model has 82,596 trainable parameters and takes approximately 8 seconds to train per epoch.

To compare how well the new loss function performs against traditional loss functions, we set up more CNN models equipped with MSE and CE loss with the same architecture as in Section \ref{sec:Model Architecture}, except that the models with MSE loss have 1-dimensional output. Mixture-of-experts ensembles of these models are formed in the same way. Learning rate schedule and dropout rates are changed slightly to ensure convergence and adequate regularization. For CNN models with MSE loss, the starting and minimum learning rates are still 0.01 and 0.001. The dropout rate is 40\%. For CNN models with CE loss, the starting and minimum learning rates are 0.005 and 0.0005. The dropout rate is 40\%.

\subsection{Net Asset Value and Sharpe Ratio}
\label{sec:Net Asset Value}

In this section, we compare the performance of all models in terms of their annual return and risk-adjusted return via the Sharpe ratio. We report the best performing linear regression models on their annual return when investing in the top 10 stocks daily, its corresponding annualized Sharpe Ratio (top 10 SR), the SR of investing in the bottom 10 recommended stocks daily (Bottom 10 SR), the long-short 10 stocks portfolio SR (Long-Short 10 SR), the SR of investing in the top decile and bottom decile daily (Top Decile SR and Bottom Decile SR), and the long-short decile spread portfolio SR (Long-Short Decile SR). The Sharpe Ratio (SR) is defined as
\begin{equation}
\label{eq:SR}
SR = \frac{\mathbb{E}[R_p - R_f]}{\sigma(R_p)},
\end{equation}
where $R_p$ is the daily percentage return of the portfolio, $R_f$ is the risk-free rate substituted as the 1-year Treasury-bill rate scaled to the daily rate, the expectation in the numerator is the mean across all test days, and the $\sigma$ in the denominator is the standard deviation. Sharpe Ratio assesses risk-adjusted returns, where a higher value indicates better performance per unit of risk.

\begin{table}[htp]
\centering
\begin{tabular}{ P{3cm} P{3cm} P{3cm} P{3cm}}
\toprule
\textbf{2005-2010} & OLS & Ridge & Lasso \\
\hline
Final Value & 1.30 & 1.25 & \textbf{1.99} \\
Annual Return & 4.98\% & 4.22\% & \textbf{13.60\%} \\
Top 10 SR & 0.30 & 0.28 & \textbf{0.45} \\
Bottom 10 SR & 0.49 & 0.50 & \textbf{0.40} \\
Long-Short 10 SR & -0.09 & -0.12 & \textbf{0.16} \\
\toprule
Top Decile SR & 0.38 & 0.38 & \textbf{0.56} \\
Bottom Decile SR & 0.38 & 0.37 & \textbf{0.23} \\
Long-Short Decile SR & 0.11 & 0.12 & \textbf{0.65}\\
\hhline{====}
\textbf{2019-2024} & OLS & Ridge & Lasso \\
\hline
Final Value & \textbf{6.02} & 5.58 & 5.98 \\
Annual Return & \textbf{40.16\%} & 38.17\% & 39.98\% \\
Top 10 SR & 0.82 & 0.80 & \textbf{0.87} \\
Bottom 10 SR & 0.95 & 0.85 & \textbf{0.6}0 \\
Long-Short 10 SR & -0.03 & 0.03 & \textbf{0.2} \\
\toprule
Top Decile SR & 0.50 & 0.49 & \textbf{0.54} \\
Bottom Decile SR & 0.74 & 0.74 & \textbf{0.6}2 \\
Long-Short Decile SR & -0.22 & -0.23 & \textbf{-0.04} \\
\bottomrule
\end{tabular}
\caption{Linear regression results}
\label{tab:linear regression}
\end{table}

For linear regression methods, in the 2005-2010 period, the set \{3-day momentum, shares traded\} achieves the best result; the set containing all 28 features leads to the best result for the 2019-2024 period. This demonstrates the shift towards requiring more factors to achieve high returns in more recent years. See Table \ref{tab:linear regression} for the detailed results of linear regression methods, where we assume the starting investment amount is 1, and the final value is calculated by investing in the top 10 stocks the model recommends daily. To be precise, this strategy compares the top 10 recommendations for the next day with the current 10 holdings. The strategy sells the current holdings that are not in the next-day recommendations, holds the current holdings that are in the next-day recommendations, and buys the remaining recommended stocks in equal dollar amounts. Note that we assume all trades are conducted at market open and that all assets are invested at all times. We use the same trading strategy to simulate the returns of all models.

The best number in each row is bolded in Table \ref{tab:linear regression}. We see that the 2019-2024 period features significantly higher returns in the top 10 category compared to the 2006-2010 period. In the 2006-2010 period, it is interesting to observe that the lasso method with cross validation has the best SR in the top 10 category even though it does not always achieve the highest return, which aligns with our expectation that it reduces overfitting that likely occurs in OLS. However, we see the decline in overall ranking abilities of these linear methods from the 2006-2010 period to the 2019-2024 period by comparing their top, bottom decile SR, and the SR of long-short decile strategy. This justifies the use of nonlinear methods on these features in the more recent market, which we discuss next.

We train three independent machine learning models according to the setup in Sections \ref{sec:Model Architecture} and \ref{sec:Hyperparameter Setting} with different randomizations and employ two methods to combine them into the final model that we test on. The first method is a simple ensemble that uses the average of the three models' predictions. The second method is a mixture of experts with dynamic weighting according to each model's historical performance. Specifically, the weight of model $i$ is 
\begin{equation}
\label{eq:mixture of experts weights}
w_i = \frac{\exp(r_i)}{\sum_{j=1}^3 \exp(r_j)},
\end{equation}
where $r_i$ is the percentage return of model $i$ in the past 6 testing periods (or since the beginning if the model has not tested for 6 times yet). We only report on the mixture-of-experts ensembles because they consistently outperform the simple average ensembles.

Dividing the entire time span into testing periods of 20 trading days, We test the performance of the mixture-of-experts ensembles equipped with the new loss function against all other models over the 68 testing periods of 20 days from 2006-2010 and 67 testing periods from 2019-2024. We calculate the same return-related statistics as before. For attention models and CNNs with the new loss function and CE loss, the ranking of all stocks is done by comparing the score. Recall that the score is calculated by the dot product of the probability distribution $(p_1, p_2, p_3, p_4, p_5)$ output corresponding to strong sell, sell, hold, buy, and strong buy, and the weights $[-2, -1, 0, 1, 2]$, that is,
\begin{equation}
\label{eq:score}
\mathrm{score} = -2p_1 - p_2 + p_4 + 2p_5.
\end{equation}
For CNNs with MSE loss, the ranking is determined by sorting the model output.

\begin{figure}[htp]
\centering
\includegraphics[width=\linewidth]{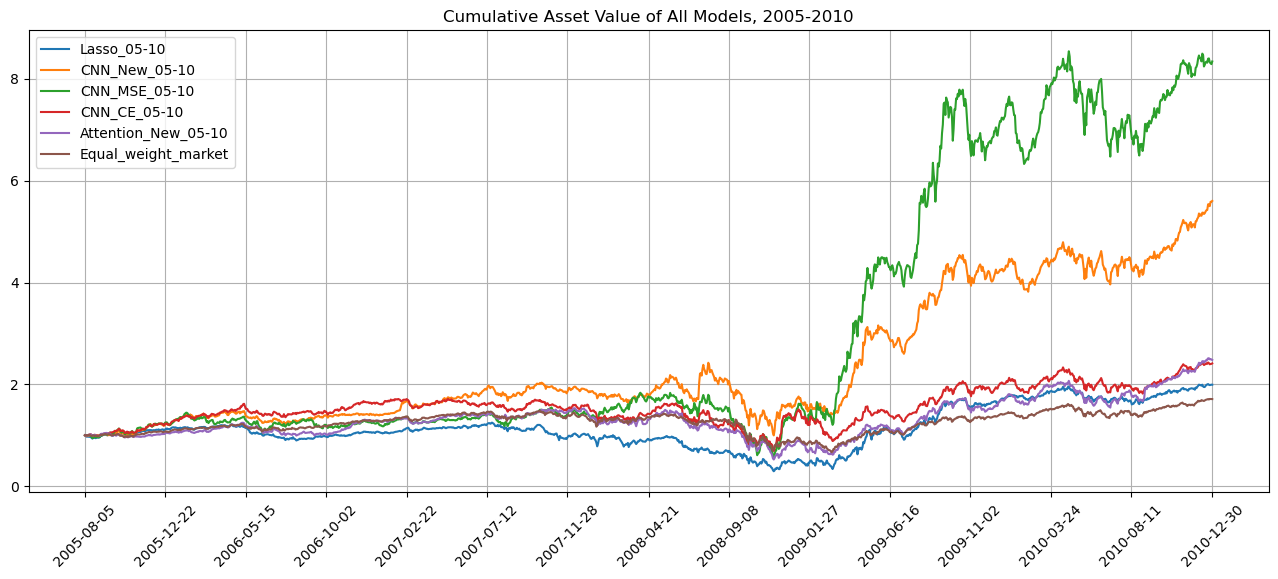}
\caption{Cumulative Asset Value of All Models, 2005-2010}
\label{fig:total asset simulation 06-10}
\end{figure}

\begin{figure}[htp]
\centering
\includegraphics[width=\linewidth]{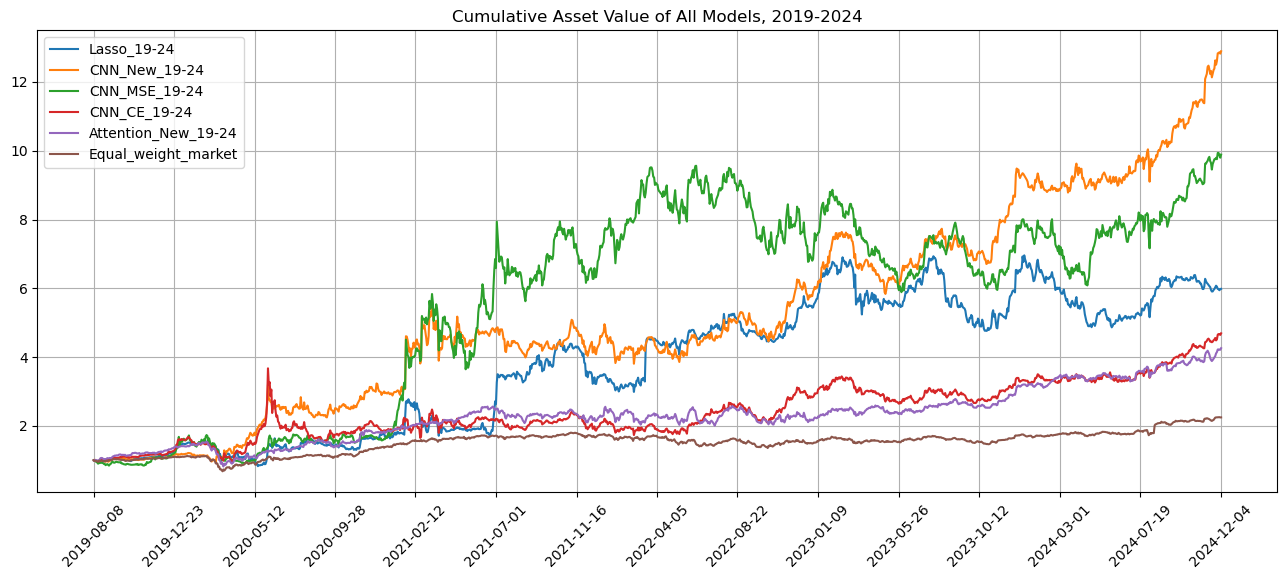}
\caption{Cumulative Asset Value of All Models, 2019-2024}
\label{fig:total asset simulation 19-24}
\end{figure}

Table \ref{tab:ML results} lists the performance of all ML models for comparison, where the attention model uses the novel loss function. We only include the mixture-of-experts ensembles as they have greater stability compared to individual models. We plot the daily portfolio value consisting of the top 10 stocks of these models in Figures \ref{fig:total asset simulation 06-10} and \ref{fig:total asset simulation 19-24}.

\begin{table}[htp]
\centering
\begin{tabular}{ P{3cm} P{2cm} P{2cm} P{2cm} P{2cm}}
\toprule
\textbf{2005-2010} & CNN\_MSE & CNN\_CE & Attention & CNN\_New \\
\hline
Final Value & \textbf{8.34} & 2.41 & 2.5 & 5.6 \\
Annual Return & \textbf{48.14\%} & 17.70\% & 18.50\% & 37.61\% \\
Top 10 SR & 0.97 & 0.56 & 0.58 & \textbf{0.97} \\
Bottom 10 SR & \textbf{0.33} & 0.56 & 0.48 & 0.34 \\
Long-Short 10 SR & 0.97 & -0.26 & 0.15 & \textbf{1.03} \\
\toprule
Top Decile SR & 0.60 & 0.49 & 0.41 & \textbf{0.64} \\
Bottom Decile SR & \textbf{0.24} & 0.48 & 0.30 & 0.35 \\
Long-Short Decile SR & \textbf{0.80} & -0.19 & 0.21 & 0.31 \\
\hhline{=====}
\textbf{2019-2024} & CNN\_MSE & CNN\_CE & Attention & CNN\_New \\
\hline
Final Value & 9.89 & 4.69 & 4.26 & \textbf{12.89} \\
Annual Return & 53.87\% & 33.73\% & 31.33\% & \textbf{61.73\%} \\
Top 10 SR & 1.00 & 0.79 & 0.97 & \textbf{1.18} \\
Bottom 10 SR & 0.85 & 0.54 & \textbf{0.21} & 0.79 \\
Long-Short 10 SR & 0.23 & 0.20 & \textbf{0.82} & 0.26 \\
\toprule
Top Decile SR & \textbf{0.86} & 0.65 & 0.73 & 0.85 \\
Bottom Decile SR & 0.80 & 0.75 & \textbf{0.47} & 0.73 \\
Long-Short Decile SR & 0.33 & -0.17 & \textbf{0.48} & 0.13 \\
\bottomrule
\end{tabular}
\caption{ML results}
\label{tab:ML results}
\end{table}

We see that all models outperform the equal-weighted market in both figures. Even though their final portfolio values differ, all ML models shown in the figures exhibit continuous growth trend throughout the approximately five years in the two periods despite being under the pressure of the 2008 financial crisis and 2020 COVID pandemic. This can be attributed to the learning rate reset in the second testing period and onward. In the first training of ML models, we gradually decrease the learning rate to ensure convergence and stability. When retraining the models in later testing periods, we reset the learning rate to its initial rate so that the models can explore the parameter space for potentially more suitable local minima. The ML models equipped with the novel loss function in both time periods are the most promising and adapt to changing market conditions very well. Their top 10 SRs are the highest compared to the others and indicate the best risk-adjusted performance.

In Figure \ref{fig:total asset simulation 06-10}, the MSE loss actually outperforms the novel loss function in terms of annual return, but lags slightly behind on SR. This points to the argument that MSE loss may have been effective in the past but has become overused in the current market, and the market can get more and more efficient with respect to it. On the other hand, the new loss function has not been widely used in the market, so it is able to achieve the best SR in both periods. It is also interesting in the 2019-2024 period that the attention model is the best in long-short strategies while slightly behind in long strategies. Comparing the annual return of the attention models and other models, we clearly see the former exhibits lesser variability and therefore has the potential to be adopted and improved in the actual market in the future.

There is a trade-off in the definition of the novel loss function \eqref{eq:loss function}. The greatest contribution to the novel loss during training lies in stocks whose next day absolute percentage change is large. Attempting to minimize the novel loss function, the model first learns to detect the patterns in each stock's input features that are closely related to impending breakouts or breakdowns. Then the model needs to classify those stocks into either Strong Buy or Strong Sell, but this task is often too difficult to do reliably. Therefore, stocks with high movement potentials are assigned a probability distribution that is heavier on the tails (Strong Sell and Strong Buy) and lighter in the middle (Sell, Hold, Buy). For stocks that do not seem to encounter significant movement or are difficult to predict, the best distribution to assign is heaviest in the middle since the original data distribution is the densest in the middle, leading to a score close to 0. As a result, in the later rows of Table \ref{tab:ML results}, the novel loss function can sometimes lag behind the other methods in overall ranking ability (Top Decile SR, Bottom Decile SR, Long-Short Decile SR) in the 2005-2010 period. However, in the more recent 2019-2024 period, the attention model is able to achieve a balance between seeking top growth opportunities and overall ranking abilities. Lastly, both the top/bottom 10 stocks and top/bottom decile portfolios have positive returns. This may be due to the overall upward trend in the market over the entire period which we point out in Section \ref{sec:First Step of Loss Function - Daily Return} and the small number of features that we consider.

\subsection{Feature Importance Analysis}
\label{sec:Feature Importance}

\begin{figure}[htp]
    \centering
    \begin{minipage}{0.47\textwidth}
        \centering
        \includegraphics[width=\textwidth]{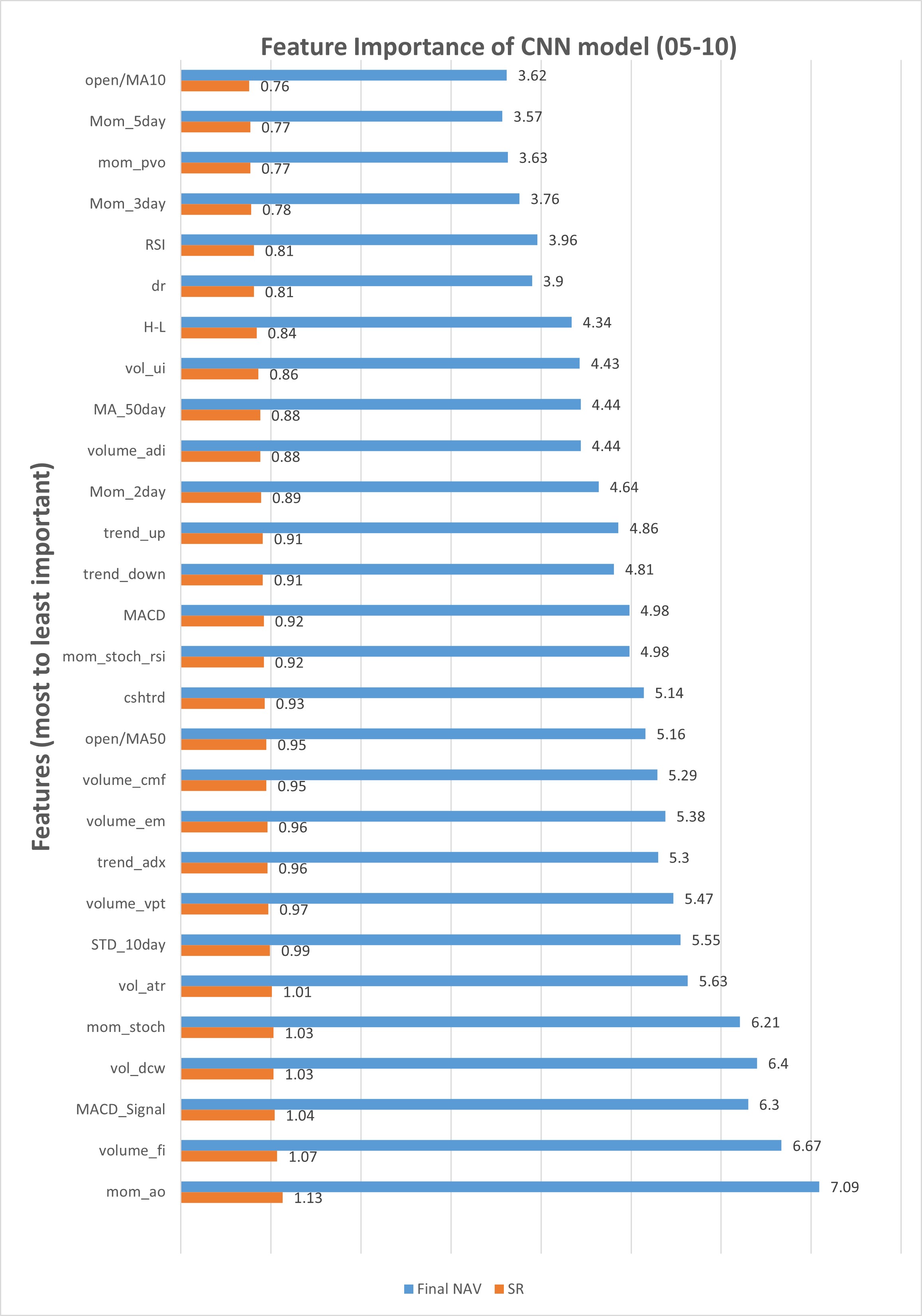}
        \caption{Feature importance of CNN model (2005-2010)}
        \label{fig:feature importance CNN 2005-2010}
        \includegraphics[width=\textwidth]{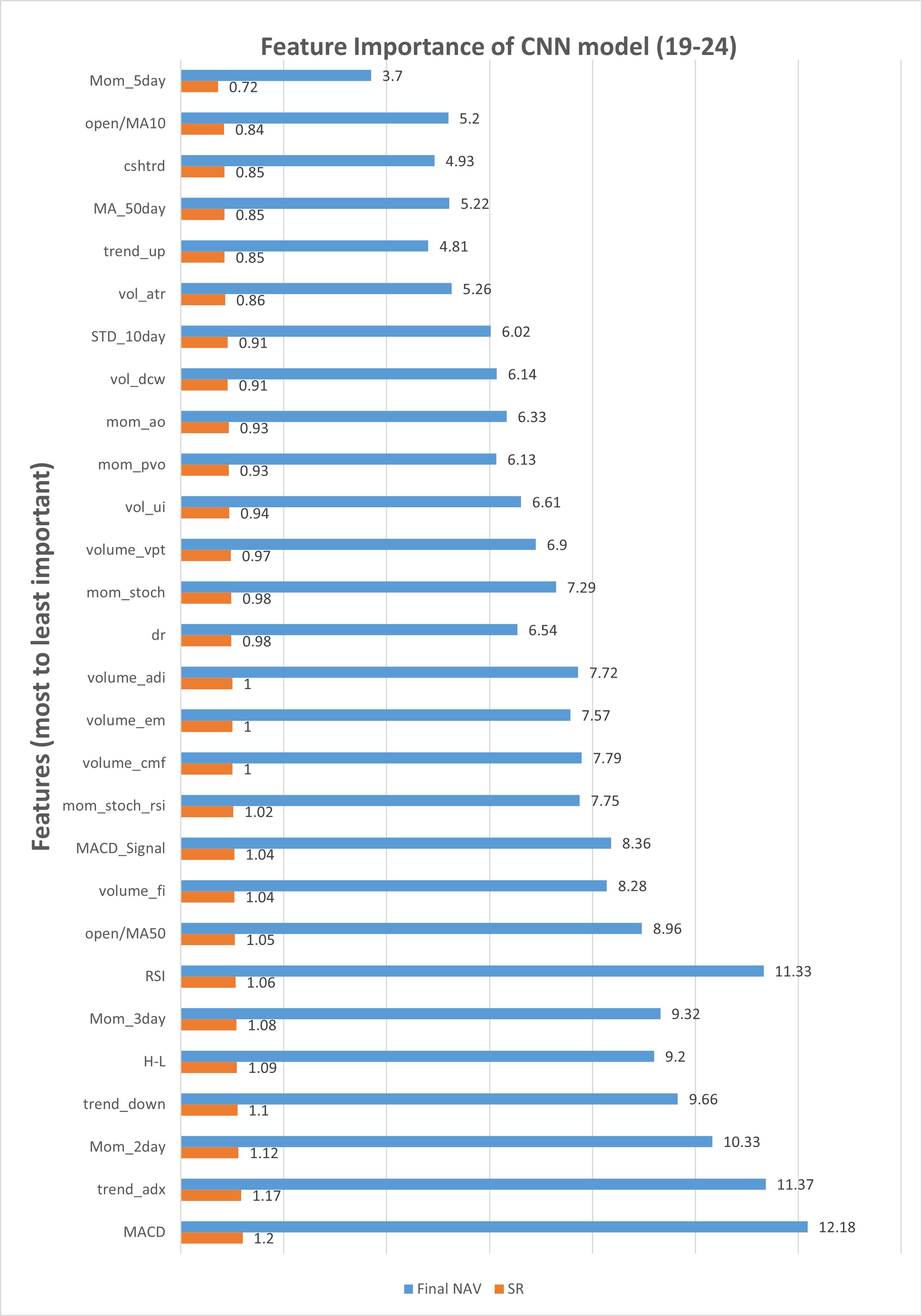}
        \caption{Feature importance of CNN model (2019-2024)}
        \label{fig:feature importance CNN 2019-2024}
    \end{minipage}\hfill
    \begin{minipage}{0.47\textwidth}
        \centering
        \includegraphics[width=\textwidth]{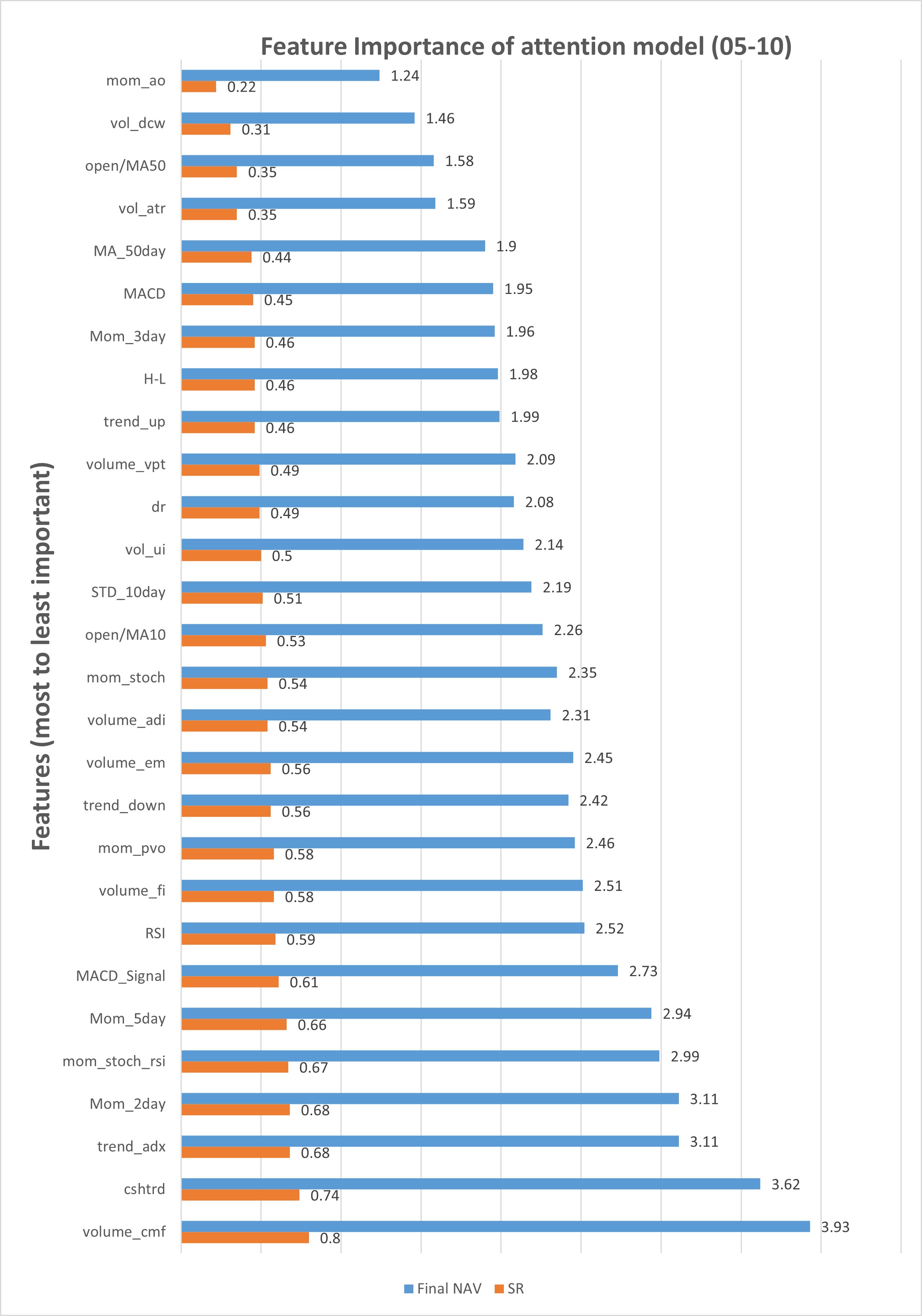}
        \caption{Feature importance of attention model (2005-2010)}
        \label{fig:feature importance attention 2005-2010}
        \includegraphics[width=\textwidth]{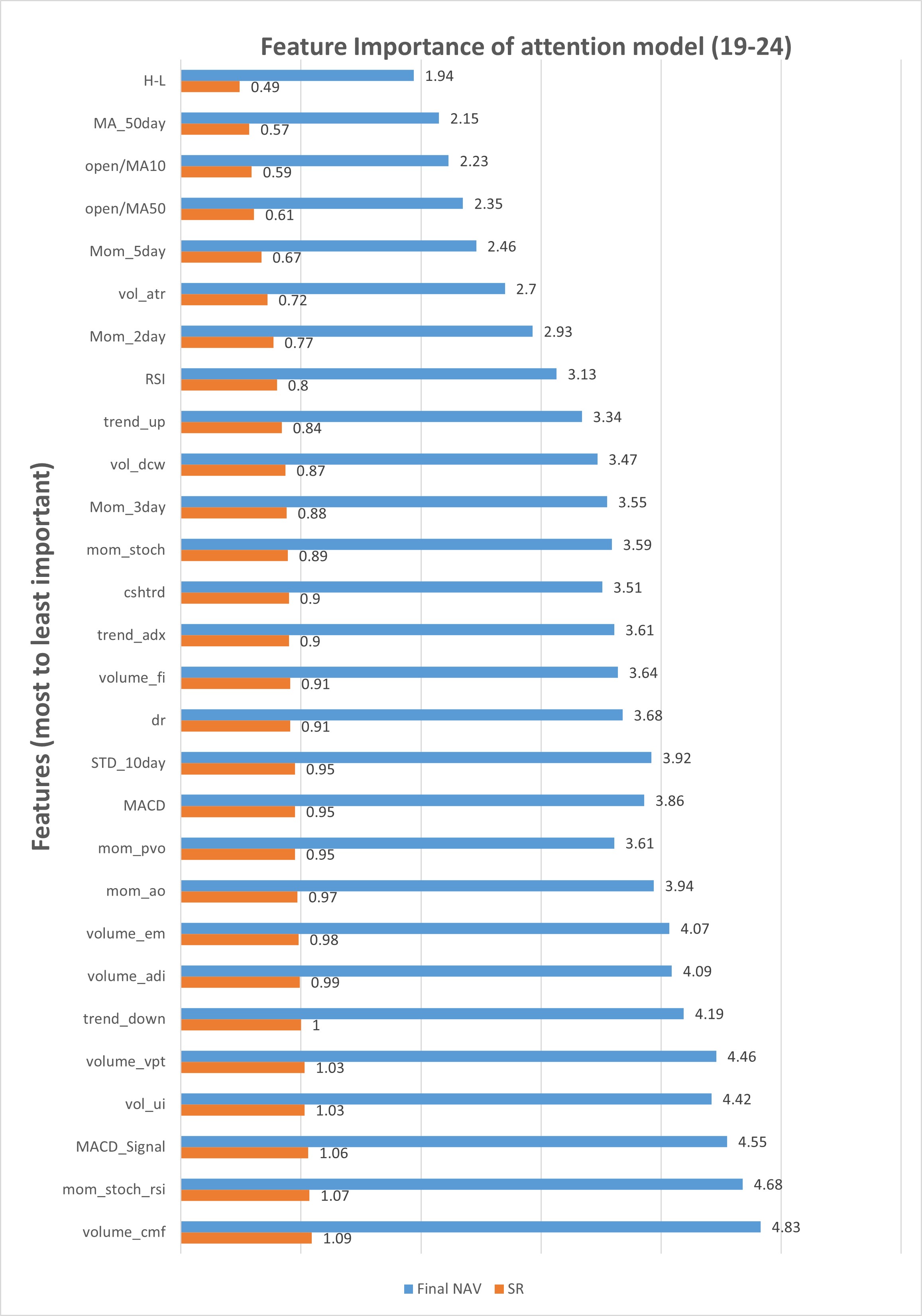}
        \caption{Feature importance of attention model (2019-2024)}
        \label{fig:feature importance attention 2019-2024}
    \end{minipage}
\end{figure}

We select the CNN and attention models in the 2005-2010 and 2019-2024 periods and analyze which features are most important in finding top growth opportunities. For each of the 28 features, we calculate the impact to the final net asset value and Sharpe Ratio by setting this feature to zero in all testing days and re-simulate keeping all else unchanged. By ranking the resulting SR from low to high absent the said feature, we obtain an ordering of all features from most to least important. Figures \ref{fig:feature importance CNN 2005-2010} and \ref{fig:feature importance attention 2005-2010} show the feature ranking in the 2005-2010 period. Figures \ref{fig:feature importance CNN 2019-2024} and \ref{fig:feature importance attention 2019-2024} show the feature ranking in the 2019-2024 period. 

Most features, when removed, negatively affect the SR; the remaining few features increase the SR when they are set to zero, so they are considered distracting features for the model. Across all four graphs, one or more short-term momentum features (1-day, 2-day, 3-day, 5-day) and moving average related features play important roles in making next-day decisions. On the other hand, many more technical features computed from the TA library tend to impact the model performance less. Given that the small number of input features we have and our short-term objective, we observe that ML models prefer simpler features and therefore may not be able to utilize the complex structure of attention to its fullest potential. This may be the reason for why attention models in this particular setting do not perform as well as the comparatively simpler CNN models in finding top growth opportunities. The amount of non-linearity introduced by convolutions on the time dimension alone is sufficient for short-term decisions.

\subsection{$t$-test and drawdown}
\label{sec:t-test}

Tables \ref{tab:compare_05-10} and \ref{tab:compare_19-24} present some performance metrics of all strategies that select the top 10 stocks during the periods 2005-2010 and 2019-2024, respectively. The key metrics include $t$-value, $p$-value, Maximum Drawdown (MD), Maximum Drawdown Duration (MDD), and annual return with SR for reference.

\begin{table}[htp]
\centering
\begin{tabular}{lrrrrrr}
\toprule
Strategy (05-10) &  Return &  $t$-value & $p$-value &  SR &  MD &  MDD \\
\midrule
CNN\_New &             37.61\% &   1.87 &   0.061 &        0.97 &       -58.86\% &                    185 \\
Attention & 18.50\% & 0.67 & 0.500 & 0.58 & -65.83\% & 441 \\
CNN\_MSE &             48.14\% &   \textbf{1.96} &   \textbf{0.050} &        0.97 &       -69.90\% &                    422 \\
CNN\_CE &             17.70\% &   0.52 &   0.604 &        0.56 &       -59.48\% &                    580 \\
Lasso &             13.60\% &   0.25 &   0.804 &        0.45 &       -76.58\% &                    516 \\
Market &             10.45\% &      - &      - &        0.41 &       -55.03\% &                    607 \\
\bottomrule
\end{tabular}
\caption{Strategies that long top 10 stocks in 2005-2010}
\label{tab:compare_05-10}
\end{table}

\begin{table}[htp]
\centering
\begin{tabular}{lrrrrrr}
\toprule
            Strategy (19-24) &  Return &  $t$-value &  $p$-value &  SR &  MD &  MDD \\
\midrule
       CNN\_New &             61.73\% &   \textbf{2.29} &   \textbf{0.022} &        1.18 &       -30.99\% &                    422 \\
       Attention & 31.33\% & 1.26 & 0.207 & 0.97 & -46.38\% & 290 \\
       CNN\_MSE &             53.87\% &   1.23 &   0.218 &        1.00 &       -47.50\% &                    604 \\
        CNN\_CE &             33.73\% &   0.94 &   0.348 &        0.79 &       -58.37\% &                   1045 \\
         Lasso &             39.98\% &   1.03 &   0.305 &        0.87 &       -46.43\% &                    234 \\
 Market &             16.43\% &      - &      - &        0.68 &       -39.62\% &                    598 \\
\bottomrule
\end{tabular}
\caption{Strategies that long top 10 stocks in 2019-2024}
\label{tab:compare_19-24}
\end{table}

We perform $t$-tests on each model's return against the equal-weighted market benchmark. A higher $t$-value and a lower $p$-value indicate greater statistical significance of returns. \citet{chen2021open} conducts statistical analysis of past cross-sectional asset pricing methods using a massive number of characteristics and regards $t$-stats greater than 1.96 as a key indicator of statistical significance. We follow the same standard and bold the statistically significant $t$ and $p$-values in Tables \ref{tab:compare_05-10} and \ref{tab:compare_19-24}. Notably, the CNN model with the new loss function in 2019-2024 has a $t$-value of 2.29 and a $p$-value of 0.022, suggesting strong statistical significance. In contrast, other strategies in 2019-2024 are far from having statistically significant performance. The strategies in 2005-2010, except for the CNN models with MSE and the new loss, do not demonstrate anything close to statistically significant performance. The MSE loss shows statistical significance, and the new loss function comes close.

Maximum Drawdown (MD) represents the largest peak-to-trough loss, indicating the level of downside risk. The ML-based strategies show varying levels of risk, with MSE loss and CE loss experiencing the most significant drawdowns of -69.90\% and -58.37\% in 2005-2010 and 2019-2024, respectively. Note that in the 2019-2024 period, the CNN model with the new loss function has the lowest MD in all models that we compare and is the only one that outperforms the market benchmark with respect to it. The new loss function has the smallest drawdown values -58.86\% in 2005-2010 and -30.99\% in 2019-2024, suggesting that our new approach has better capital preservation. Maximum Drawdown Duration (MDD) measures the longest period required to recover from a drawdown. Our new approach recovers from the 2008 economic crisis very quickly in just 185 days and outpaces the other methods with the second best needing 422 days. CE loss in 2019-2024 exhibits the longest recovery duration (1045 days), while our new loss recovers the fastest among the ML models. The recovery period of the new loss is longer than lasso regression, but the lasso method has much worse return, SR, and $t$-value. Overall, our new loss function demonstrates superior performance in terms of return and risk-adjusted return while maintaining a relatively moderate drawdown.

\subsection{Strategy Capacity}
\label{sec:Strategy Capacity}

\begin{figure}[htp]
\centering
\includegraphics[width=\linewidth]{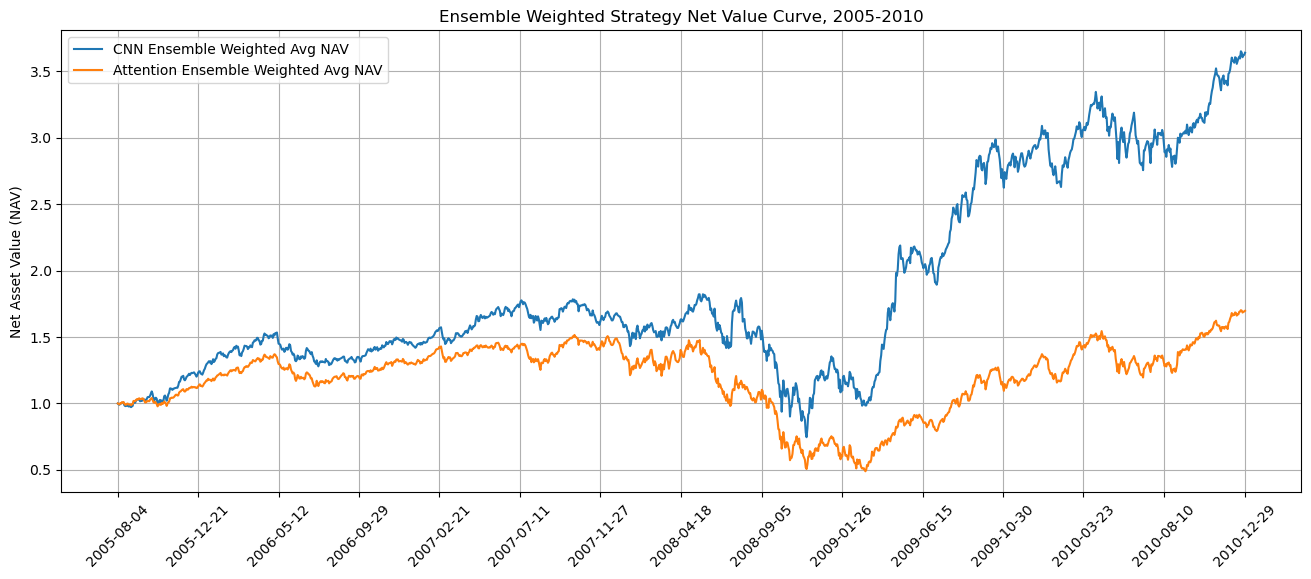}
\caption{Net Asset Values of Combining Mixture-of-Experts CNN Ensembles with New Loss, 2005-2010}
\label{fig:NAV_05-10}
\end{figure}

\begin{figure}[htp]
\centering
\includegraphics[width=\linewidth]{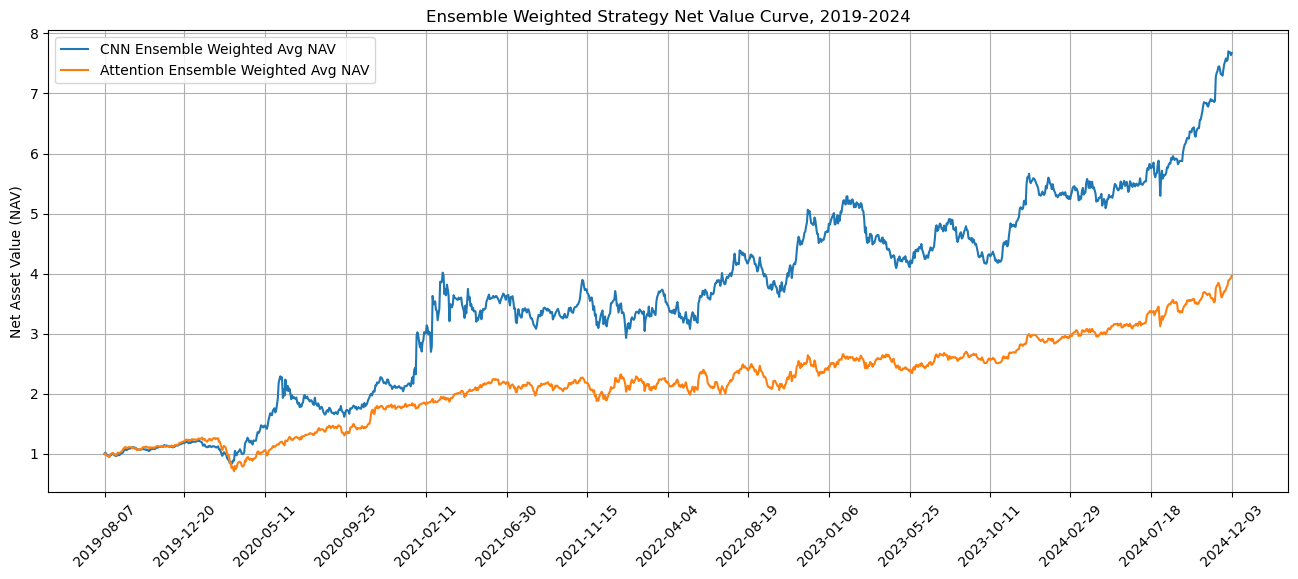}
\caption{Net Asset Values of Combining Mixture-of-Experts CNN Ensembles with New Loss, 2019-2024}
\label{fig:NAV_19-24}
\end{figure}

We also simulate an equal-weighted strategy containing multiple mixture-of-experts ML models that we have trained with the new loss function to both increase the portfolio holding capacity by multiple times and to avoid any possible bias from an individual ensemble. Figures \ref{fig:NAV_05-10} and \ref{fig:NAV_19-24} show the net asset value curves of combined ensembles for the 2005-2010 period and the 2019-2024 period, respectively. The SRs of these integrated strategies (0.79 for CNN, 0.38 for attention in 2005-2010 and 1.08 for CNN and 0.98 for attention in 2019-2024) are marginally less than the best model we have ever generated, showing that our method is remarkably consistent despite using different randomization in the training process.

\section{Conclusions and Future Directions}
\label{sec:Conclusions}

In this study, we build and test an efficient daily trading system utilizing a novel loss function to pursue top growth opportunities. We add the sector embedding cross-sectionally to adjust each feature's impact in different sectors (and optionally using multi-head attention first), perform 1D convolutions in the time dimension, use fully connected layers to reach a probability distribution output, and train the model with the novel return-weighted loss function. In each retraining period, we reset the learning rate to allow the models to adjust for evolving market conditions. We utilize mixture-of-experts ensemble to combine individual models and increase stability. In terms of return and risk-adjusted return (SR), our new loss function outperforms other traditional loss functions, especially in the later period of 2019-2024. The new loss function also shows a smaller maximum drawdown and faster recovery time comparatively, even under the difficult 2008 financial crisis and 2020 COVID pandemic. Statistical analysis establishes that the superiority of the new loss function is unlikely by chance in the more recent period, and that it is only behind MSE loss marginally in the earlier period. We further increase the capacity of the model by integrating several independently trained mixture-of-experts ensembles with different randomizations. This removes the potential bias from only reporting the best ensemble model.

In the analysis, we observe that none of the models shows significant downside in short positions. We believe this is largely due to the very small number of features that we consider and the overall upward trend in the US market. Note that we do not provide the models with fundamental characteristics (such as PE, EPS, and even market capitalization) that speak a lot about each company's intrinsic value, nor do we incorporate any textual input such as news releases and 10-K, 10-Q reports. These considerations are to be explored in future studies, and we believe this will improve the model performance further when equipped with the novel loss function.

Another limitation of the strategy is that its daily returns have high variability. The CNN model in the 2019-2024 period has an over 60\% annual return but its Sharpe Ratio is only 1.18. This is evidently due to high standard deviation in daily portfolio values. To alleviate this problem and further improve the Sharpe ratio, we propose future work that includes considering more diversification (such as in the form of Section \ref{sec:Strategy Capacity}) and improved asset allocation instead of equal allocation in the current setting. Since our model identifies stocks with immediate large movement, it also has the potential to be integrated into straddle and reverse iron condor strategies, for example. With the proposal of our new loss function guiding the training process, this study provides new ideas towards incorporating actual financial impact into model construction and increasing the utility of machine learning models in financial markets.

\bmhead{Acknowledgements}

The first author is grateful for Tyler Maunu for inspiring conversations about the contents of the paper. The first author also thanks Martin Molina Fructuoso for helpful discussions in the early stage of the project. This research did not receive any specific grant from funding agencies in the public, commercial, or not-for-profit sectors.

\begin{appendices}
\section{Technical Features}
\begin{longtable}{ P{0.2\linewidth} P{0.8\linewidth}}
\toprule
Momentum Indicators & Description \\
\hline
Stochastic RSI & Combines RSI and stochastic oscillator to identify overbought or oversold conditions \\
Stochastic Oscillator & Measures the current price vs. price range to identify potential reversal points \\
Awesome Oscillator & Calculates the difference between two simple moving averages to affirm trends or identify reversals \\
Percentage Volume Oscillator & Measures the percentage difference between two volume-based moving averages to indicate changes in volume momentum \\
Kaufman's Adaptive Moving Average & Adjusts to market noise or volatility through price swings for identifying trends and filtering price movements \\
Williams \%R & Reflects the level of price relative to the highest high for a period to indicate overbought or oversold conditions \\
\toprule
Volume Indicators & Description \\
\hline
Accumulation Distribution Index & Combines price and volume to assess whether a stock is being accumulated or distributed \\
Ease of Movement & Relates the rate of price change to volume to indicate the ease with which prices are moving \\
Force Index & Combines price change and volume to assess the force behind a price movement \\
Chaikin Money Flow & Measures the volume-weighted average of accumulation and distribution over a specified period to indicate buying or selling pressure \\
Volume-Price Trend & Combines price trend and volume to show the strength of price movements \\
\toprule
Volatility Indicators & Description \\
\hline
Average True Range & Measures market volatility by decomposing the entire range of an asset price over a given period \\
Bollinger Bands High & Plots standard deviation levels above a moving average to indicate high volatility and potential overbought conditions \\
Donchian Channel Width & Calculates the difference between the highest high and the lowest low over a specified period to indicate volatility \\
Ulcer Index & Measures downside risk by quantifying the depth and duration of price declines \\
\toprule
Trend Indicators & Description \\
\hline
Average Directional Index & Quantifies the strength of a trend without considering its direction \\
Aroon Up & Measures the time since the highest high within a specified period to indicate the strength of an uptrend \\
Aroon Down & Measures the time since the lowest low within a specified period to indicate the strength of a downtrend \\
Ichimoku Leading Span A & Part of the Ichimoku Cloud system, represents a midpoint between two averages, used to identify support and resistance levels\\
\bottomrule
\caption{Brief Description of All Technical Features}
\label{tab:all features}
\end{longtable}
\end{appendices}

\section*{Declarations}
\textbf{Funding} The authors declare that no funds, grants, or other support were received during the preparation of this manuscript.

\vspace{\baselineskip}

\noindent\textbf{Competing Interests} The authors have no relevant financial or non-financial interests to disclose.

\vspace{\baselineskip}

\noindent\textbf{Data Availability} The data used in this study can be obtained from the public online repository \href{https://data.mendeley.com/datasets/czwwfgcgz7/1}{Mendeley Data}.

\bibliography{refs}

\end{document}